\documentclass[letterpaper, 10 pt, conference]{ieeeconf}  

\IEEEoverridecommandlockouts                              

\usepackage{textcomp}
\usepackage{multirow}
\usepackage{hyperref}
\usepackage{cite}
\usepackage[pdftex]{graphicx}
\graphicspath{{images/}}
\DeclareGraphicsExtensions{.pdf,.jpeg,.png}
\usepackage[cmex10]{amsmath}

\usepackage{array}

\usepackage[caption=false,font=normalsize,labelfont=sf,textfont=sf]{subfig}

\usepackage{color}
\usepackage{xcolor}

\usepackage[normalem]{ulem}

\usepackage{amsfonts}
\usepackage{amssymb}

\title{\LARGE \bf
Omnipotent Virtual Giant for Remote Human--Swarm Interaction
}

\author{Inmo~Jang,~%
        Junyan~Hu,~%
        Farshad~Arvin,~%
        Joaquin~Carrasco,~%
        and~Barry~Lennox
\thanks{*This work was supported by EPSRC RAIN project No. EP/R026084/1}
\thanks{I. Jang, F. Arvin, J. Carrasco, and B. Lennox are with Robotics for Extreme Environment Group, School of Electrical \& Electronic Engineering, The University of Manchester, M13 9PL, United Kingdom
(e-mail: \tt\small\{inmo.jang\},\{farshad.arvin\}@manchester.ac.uk). }
\thanks{J. Hu is with Control Systems Group, School of Electrical \& Electronic Engineering, The University of Manchester, M13 9PL, United Kingdom}
}

\begin{document}

\maketitle
\thispagestyle{empty}
\pagestyle{empty}

\begin{abstract}
This paper proposes an intuitive human-swarm interaction framework inspired by our childhood memory in which we interacted with living ants by changing their positions and environments as if we were omnipotent relative to the ants.
In virtual reality, analogously, we can be a super-powered virtual giant who can supervise a swarm of mobile robots in a vast and remote environment by flying over or resizing the world, and coordinate them by picking and placing a robot or creating virtual walls.    
This work implements this idea by using \emph{Virtual Reality} along with \emph{Leap Motion}, which is then validated by proof-of-concept experiments using real and virtual mobile robots in mixed reality.  
We conduct a usability analysis to quantify the effectiveness of the overall system as well as the individual interfaces proposed in this work.  
The results revealed that the proposed method is intuitive and feasible for interaction with swarm robots, but may require appropriate training for the new end-user interface device. 
\end{abstract}

\section{Introduction}\label{sec:introduction}

Swarm robotics~\cite{hamann2018swarm} is one of the promising robotic solutions for complex and dynamic tasks thanks to its inherent system-level robustness from the large cardinality.
Swarm robotics researches mostly consider small and individually incapable robots, for example, Kilobots\cite{Trianni2018} and MONAs\cite{Arvin2018}, but the resultant technologies and knowledge can be also transferable to a swarm of individually capable robots (e.g. legged robots), which will be deployed for important and safety-critical missions in extreme environemtns such as nuclear facility inspection.  

%
\emph{Human-Swarm Interaction} (HSI) is relatively a new research area that ``aims at investigating techniques and methods suitable for interaction and cooperation between humans and robot swarms''\cite{Nagi2015}. 
One of the main differences of HSI from typical \emph{Human-Robot Interaction} (HRI) is that a large number of robots, due to swarm properties, must be involved efficiently, otherwise a human operator can be easily overwhelmed by enormous workload for control and situational awareness.
In addition, it is highly expected that swarm robots are controlled by decentralised local decision-making algorithms \cite{Choi2009, Jang2018a, Jang2018d}, which generate a desired emergent group behaviour. Therefore, HSI should be \emph{synergistic} with such self-organised behaviours by having interfaces of not only individual-level teleoperation but also subgroup-level and mission-level interactions.  
Furthermore, in practice, e.g. in an extreme environment, swarm robots will be deployed to a mission arena beyond the line-of-sight of a human operator. 
Therefore, considering such possible scenarios, HSI should be differently addressed than typical HRI.

\begin{figure}[t]
\centering
\includegraphics[width=0.99\linewidth]{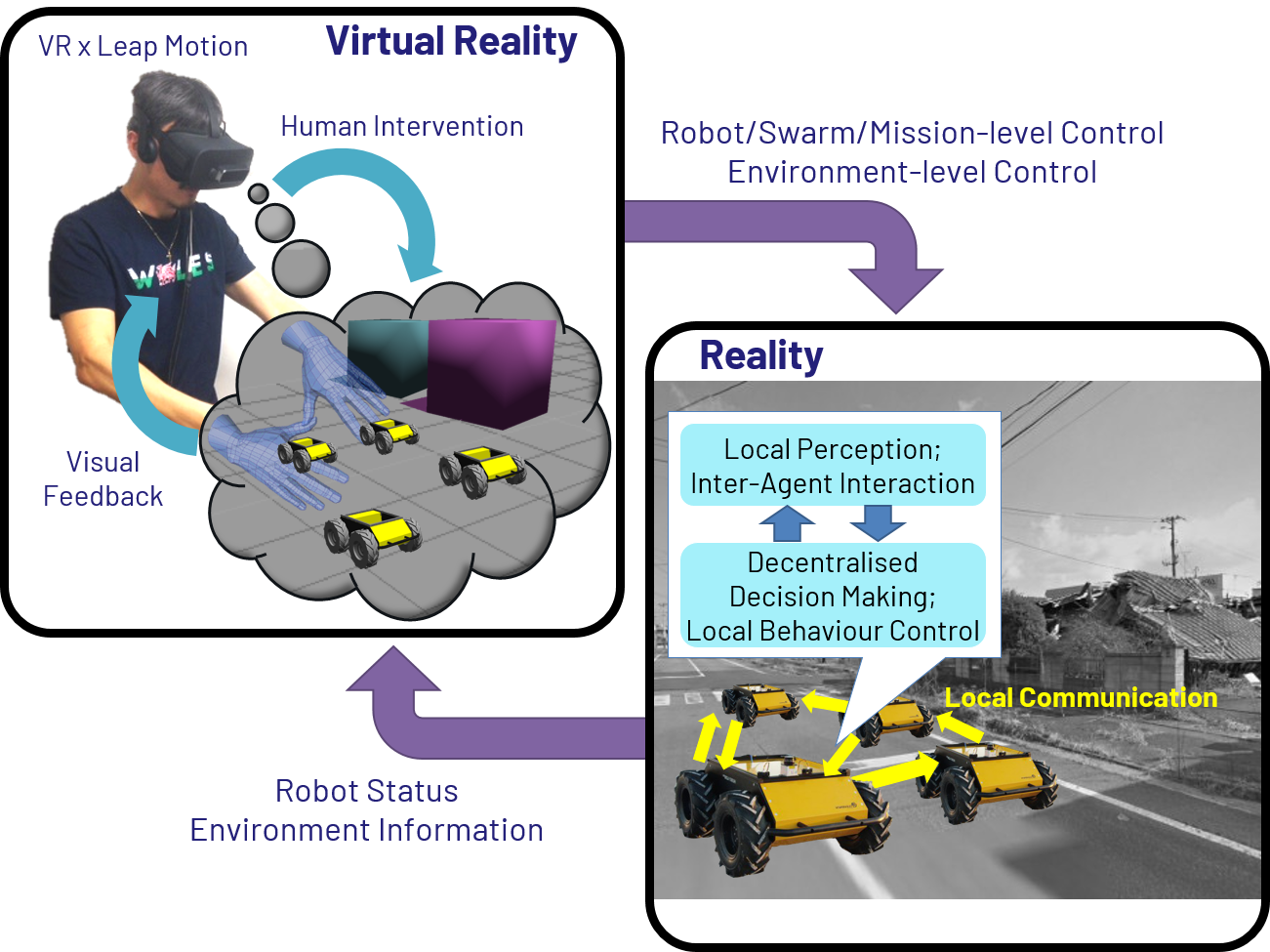}
\caption{The system architecture of the proposed Human-Swarm Interaction using \emph{Omnipotent Virtual Giant}}
\label{fig:hsi_framework}
\end{figure}

In this paper, we propose an \emph{Omnipotent Virtual Giant} for HSI, which is a super-powered user avatar interacting with robot swarm via virtual reality, as shown in Fig. \ref{fig:hsi_framework}. 
This was inspired by our childhood memory in which most of us have played with living ants by relocating their positions and putting obstacles on their paths as if we were omnipotent relative to them. 
Analogously, through the omnipotent virtual giant, a human operator can directly control individual robots by picking and placing them; can alter virtual environment (e.g. creating virtual walls) to indirectly guide the robots; and can be omniscient by flying around or resizing the virtual world and supervising the entire or a subgroup of the robots. 
We implement this idea using \emph{Leap Motion} with \emph{Virtual Reality} (Sec. \ref{sec:proposed_HSI}), and validate the proposed HSI framework by using proof-of-concept real-robot experiments and by usability tests (Sec. \ref{sec:experiments}). 

%


\section{Related Work}\label{sec:literature}
This section particularly reviews existing HSI methodologies and their suitability for remote operations. 
\emph{Gesture-based} interactions have been popularly studied\cite{Nagi2014, Nagi2015, Podevijn2014, Alonso-Mora2015, Stoica2013, Gromov2016}. 
A human's body, arm, or hand gestures are recognised by \emph{Kinect} \cite{Podevijn2014, Alonso-Mora2015}, electromyography sensors \cite{Stoica2013, Gromov2016}, or onboard cameras \cite{Nagi2014,Nagi2015}, and then translated to corresponding commands to robots.   
Such gesture-based languages probably require a human operator to memorise mappings from predefined gestures to their intended commands, although some of the gestures may be intuitively used.

\emph{Augmented Reality (AR)} has been also utilised in 
\cite{Frank2017, Patel2019}. 
This method generally uses a tablet computer, which, through its rear-view camera, recognises robots and objects in the environment. Using the touchscreen, a user can control the robots shown on the screen, for example, by swipe gestures. 
In \cite{Patel2019}, an AR-based method was tested for cooperative transport tasks of multiple robots. 
However, this type of interface is only available in the proximity environment. 

\emph{Tangible} interactions can be another methodology for certain types of swarm robots. The work in \cite{LeGoc2016} presented tiny tabletop mobile robots, with which a human can interact \emph{by actually touching} them. By relocating a few of the robots, the entire robots eventually end up with different collective behaviours. 
This tangible interface inherently does not allow any interfacing error when it comes to changing of a robot's position. 
Nevertheless, apart from position modifications, it seems not straightforward to include the other interfaces.

All the aforementioned interfaces require a human operator to be within proximity of robots. 
Instead, \emph{virtual reality (VR)}-based interactions can be considered as an alternative for beyond-line-of-sight robotic operations.
In a virtual space where a human operator interacts with swarm robots, the operator is able to violate the laws of physics, 
teleporting \cite{Roldan2019} or resizing the virtual world (as will be shown in this paper) to observe the situation macroscopically. 
This may facilitate to perceive and control a large number of robots in a vast and remote environment. 
However, most of existing VR-based interfaces rely on default hand-held equipment. They would be less intuitive than using our bare hands, but also may cause considerable load on the user's arms when in use for a longer time.

\section{Methodology: Omnipotent Virtual Giant}\label{sec:proposed_HSI}

In this paper, we proposed a novel HSI framework using \emph{omnipotent virtual giant}, which is a resizable user avatar who may perceive situations macroscopically in virtual space but also can interact with swarm robots by using its bare hands, e.g. simply picking and placing them.
Technically, this concept can be implemented by integrating virtual reality (VR) and Leap Motion (LM). 
Our proposed method has both advantages of tangible interactions giving intuitiveness as well as VR-based interactions giving remote operability. 



\subsection{Preliminary}

\subsubsection{Virtual Reality}
VR is considered as one of the suitable user interfaces to interact with remote swarm robots\cite{Roldan2019}. 
On top of its advantages described in the previous section, 
VR being the main interface device can provide practical efficiency in research and development (R\&D) process. 
In general, developing user interfaces requires enormous human trials and feedback update process via numerous beta tests. 
This process can be accelerated, if VR is in use, by using simulated swarm robots in the initial phase of R\&D, which is a very important period to explore various design options within a relatively short time. 
For example, for real swarm robotic tests, it may take elongated time to prepare such a large number of robots (e.g. charging batteries), which can be avoidable when simulated robots are instead in use.
In addition, by using robot simulators (e.g. \emph{Gazebo}, \emph{V-Rep}, \emph{ARGoS}\cite{Pinciroli2011}) along with communication protocols such as \emph{rosbridge} and \emph{ROS\#}, we can construct \emph{mixed reality}\cite{Roldan2019, Whitney2018}, where real robots and simulated robots coexist, and then perform a hardware-in-the-loop test with the reduced R\&D resources (e.g. human power, time, and cost). Obviously, the final phase of R\&D should involve proper real robot tests in fields, however, thanks to VR, unnecessary efforts can be reduced over the whole development period. 

\subsubsection{Leap Motion}
LM is a vision-based hand motion capture sensor. 
Recently, performance of LM has been significantly improved in the latest SDK called \emph{Orion}\footnote{Watch this comparison video: \href{https://youtu.be/7HnfG0a6Gfg}{https://youtu.be/7HnfG0a6Gfg}}.
Particularly, when it is used along with \emph{Unity}\footnote{https://unity3d.com/}, we can exploit useful modules (e.g. Leap Motion Interaction Engine) that facilitate to interact with virtual objects using bare hands without any hand-held equipment. 
In our previous work~\cite{Jang2019}, 
hands sensed by LM are reasonably accurate and much more natural to use compared with the use of hand-held devices.

\subsection{System Overview}

The architecture of the proposed HSI framework, as illustrated in Fig. \ref{fig:hsi_framework}, consists of the following subsystems: 

\begin{itemize}
    \item \emph{Mobile robots}: Swarm robots are deployed to a remote mission area. 
    The robots are assumed to have capabilities of decentralised decision making\cite{Choi2009, Jang2018a, Jang2018d}, navigation and control (e.g. path planning, collision avoidance, low-level control, etc.) \cite{Jang2018}, remote inspection\cite{Bird2018}, manipulation\cite{martinoli-04}, and inter-agent communication. 
    They behave autonomously based on their local information and interaction with their neighbouring robots.
   
    \item \emph{Data collection from the robots and visualisation}:
    The status of the robots and the environments where they are inspecting are transmitted to the master control station, where this information is assumed to be dynamically rendered in virtual reality. 
    This communication may happen in a multi-hop fashion since the network topology of the robots is not likely to be fully connected.
    \item \emph{Interactions via an omnipotent virtual giant}: A user wearing a VR head-mounted display can perceive the remote situation through the virtual reality. 
    The user's bare hands are tracked by the LM attached on the outer surface of the VR goggle, and then rendered as the hands of the avatar in the virtual space. 
    The user avatar is resizeble to become a giant or flyable around to oversee the overall situation. The user can interact with the robots by touching them in the virtual space. 
     The details of the user interfaces currently implemented will be described in Sec. \ref{sec:proposed_interface}.
    \item \emph{User input transmission to the robots}:
    When an interaction happens in the virtual space, corresponding user inputs are sent to the real robots, and they react accordingly.
     
\end{itemize}

This work mainly focuses on the user interaction part of the system. It is assumed that all the other subsystems are provided, which are beyond the scope of this paper.

\subsection{Proposed User Interfaces}\label{sec:proposed_interface}

This section describes user interfaces we propose in this work. 
The main hand gestures used are as follows:
\begin{itemize}
	\item \emph{Pinching}: This gesture is activated when the thumb and index finger tips of a hand are spatially close as shown in Fig. \ref{fig:ui_perception}(a). 
	\emph{PinchDetector} in LM SDK facilitates this gesture. 
	
	\item \emph{Closing hand}: This is triggered when all the five fingers are fully closed as in Fig. \ref{fig:ui_perception}(b).
	When this is done, the variable \emph{GrabStrength} $\in [0,1]$ in the class \emph{Leap::Hand} of the SDK becomes one.

	\item \emph{Grasping}: This will begin if a thumb and index finger are both in contact with an object. If this is initiated, the object can be grasped. 
	One example is shown in Fig. \ref{fig:UI_individual_control}. 
	This gesture can be implemented via \emph{Leap Motion Interaction Engine}. 
	
	\item \emph{Touching}: Using an index finger, virtual buttons can be pushed as in Fig. \ref{fig:UI_menu}(a). 
\end{itemize}
Combination of the gestures is used for perception or control for swarm robots.


\subsubsection{Perception interfaces}

Given robot swarm spread in a vast arena, capabilities of overall situation awareness as well as robot-level perception are crucial for HSI. To this end, this paper proposes the following two interfaces: \emph{Resizing the world} and \emph{Flying}. 

\begin{figure}
\centering
\subfloat[]{\includegraphics[width=0.49\linewidth]{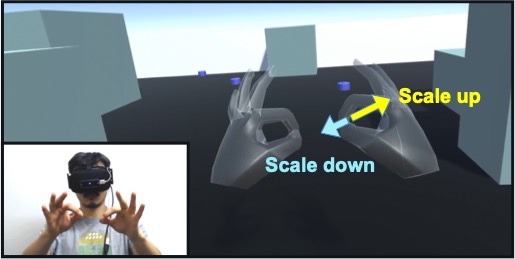}}
\hfil
\subfloat[]{\includegraphics[width=0.49\linewidth]{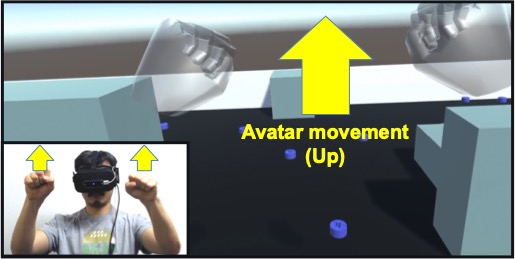}}
\hfil
\subfloat[]{\includegraphics[width=0.49\linewidth]{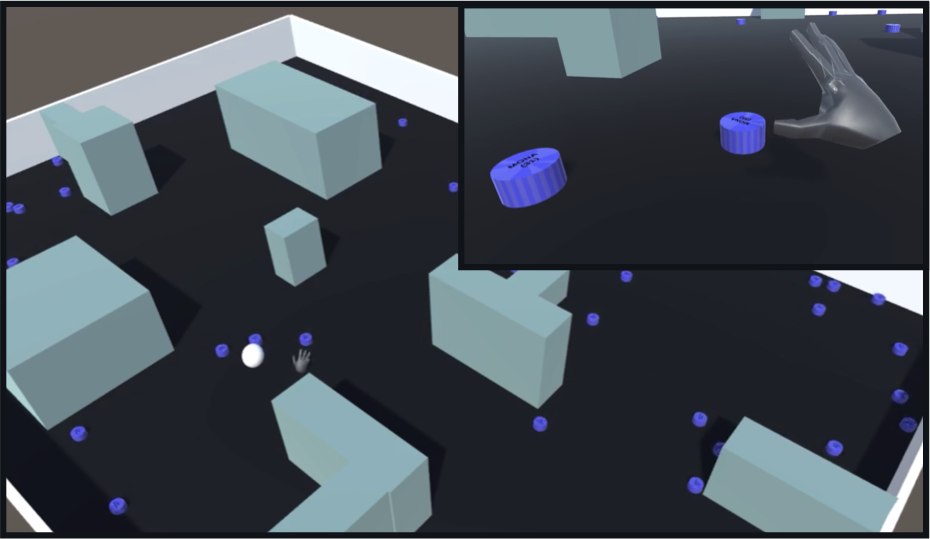}}
\hfil
\subfloat[]{\includegraphics[width=0.49\linewidth]{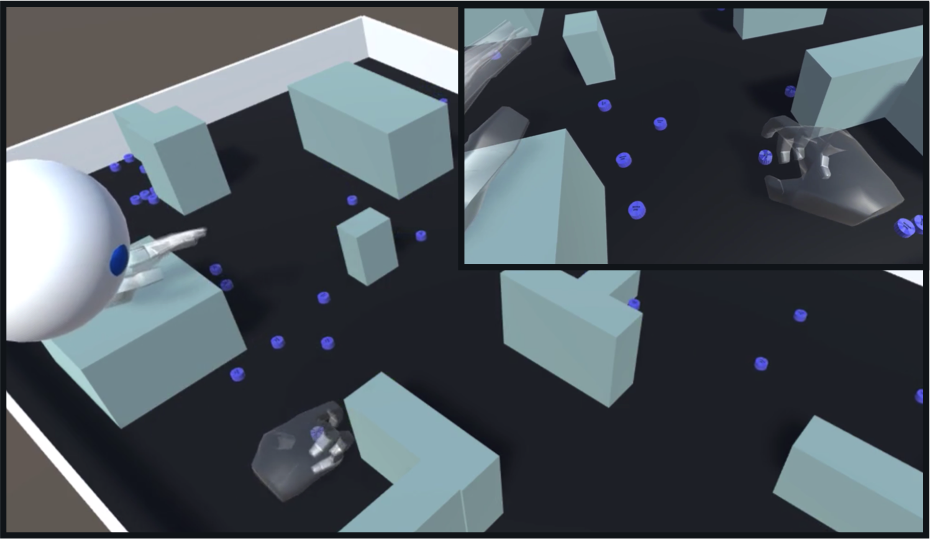}}
\hfil
\caption{Perception interfaces: (a) resizing the world; (b) flying like Superman. Using the interfaces, a user can have (c) an ordinary perception, or (d) macroscopic perception for which the avatar becomes a virtual giant. In (c) and (d), the white oval object indicates the avatar's head, and the upper-righthand subfigures show the user's view.}\label{fig:ui_perception}
\end{figure}

\textbf{Resizing the world:}
When two hands pinching are spread out or drawn together as shown in Fig. \ref{fig:ui_perception}(a), the virtual world is scaled up or down, respectively. Meanwhile, the size of the user avatar remains unchanged. 
In other words, the user avatar can be a virtual giant to oversee the situation macroscopically (Fig. \ref{fig:ui_perception}(d)) or become as small as an individual robots to scrutinize a specific area (Fig. \ref{fig:ui_perception}(c)).


\textbf{Flying like Superman:}
The user avatar basically hovers the virtual world, not being under gravity. 
Furthermore, it can even fly towards any direction by closing two hands and slightly stretching out the arms towards the same direction. 
With respect to the middle point of the two hands starting the closing-hand gesture, the relative vector of the current middle point is used as the user's intended flying direction.

\subsubsection{Control Interfaces}

User interactions to guide and control multiple robots can be summarised as the following four categories\cite{Alonso-Mora2015, Patel2019, Stoica2013}: \emph{robot-oriented}; \emph{swarm-oriented}; \emph{mission-oriented}; and \emph{environment-oriented}. 
In robot-oriented interaction, a human operator overrides an individual robot's autonomy, giving an explicit direct command, e.g. teleoperation. 
Swarm-oriented interaction uses a set of simplified degrees of freedom to control swarm robots, for example, controlling a leader robot followed by some of the other robots.  
In mission-oriented interaction, a human user provides a mission statement or plan to swarm robots as a higher-level interaction.
For swarm or mission-oriented interactions, collective autonomy or swarm intelligence takes a crucial part to achieve the desired emergent behaviour. 
Environment-oriented interaction does not affect the autonomy of any single robot, instead modifies the environments which the robots interact with, for example, by giving artificial pheromones\cite{Arvin2015}. 

In this work, we present one interface per interaction mode except mission-oriented one, which are as follows: \emph{Pick-and-Place a Robot} (for robot-oriented interaction), \emph{Multi-robot Controlling Cube} (for swarm-oriented one), and \emph{Virtual Wall} (for environment-oriented one).

\begin{figure}[t]
\includegraphics[width=0.49\linewidth]{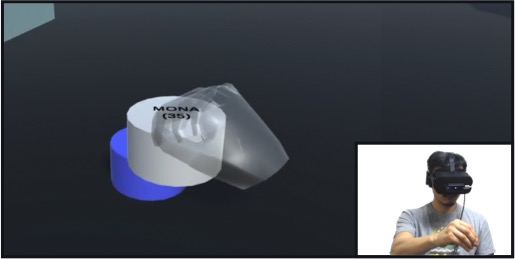}
\includegraphics[width=0.49\linewidth]{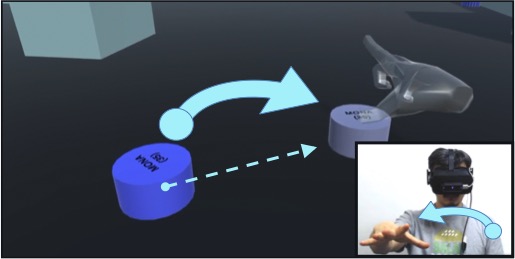}
\caption{Robot-oriented interface: picking and placing a robot}\label{fig:UI_individual_control}
\end{figure}

\textbf{Pick-and-Place a Robot:}
When the user avatar grasps a mobile robot, the robot's holographic body, which is its \emph{target-to-go} object, is picked up and detached from the robot object, as shown in Fig. \ref{fig:UI_individual_control}.
Once the target-to-go object is relocated to any position, then 
the robot moves towards it while neglecting its existing autonomy.  

\begin{figure}[t]
\centering
\subfloat[]{\includegraphics[width=0.49\linewidth]{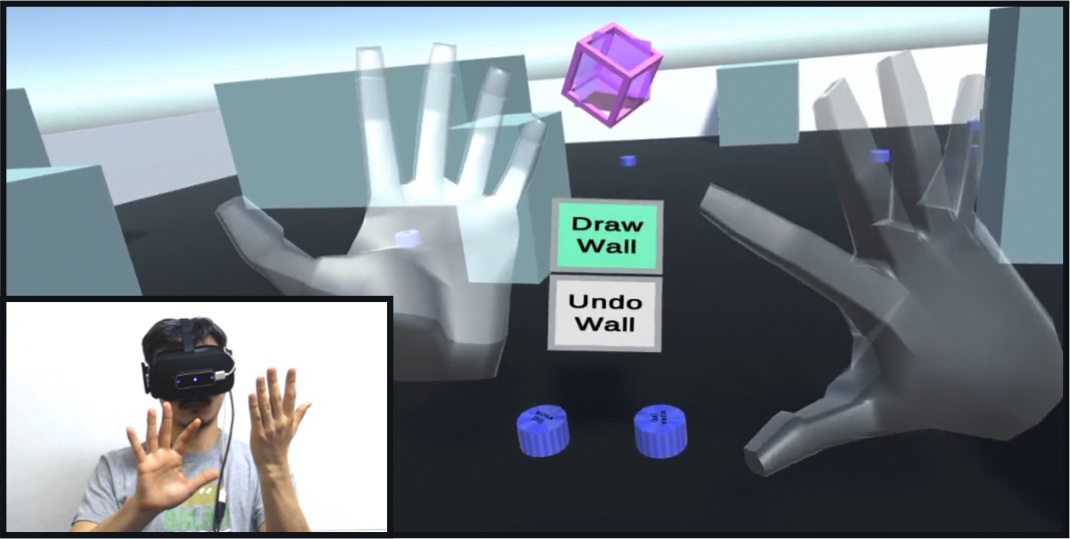}}
\hfil
\subfloat[]{\includegraphics[width=0.49\linewidth]{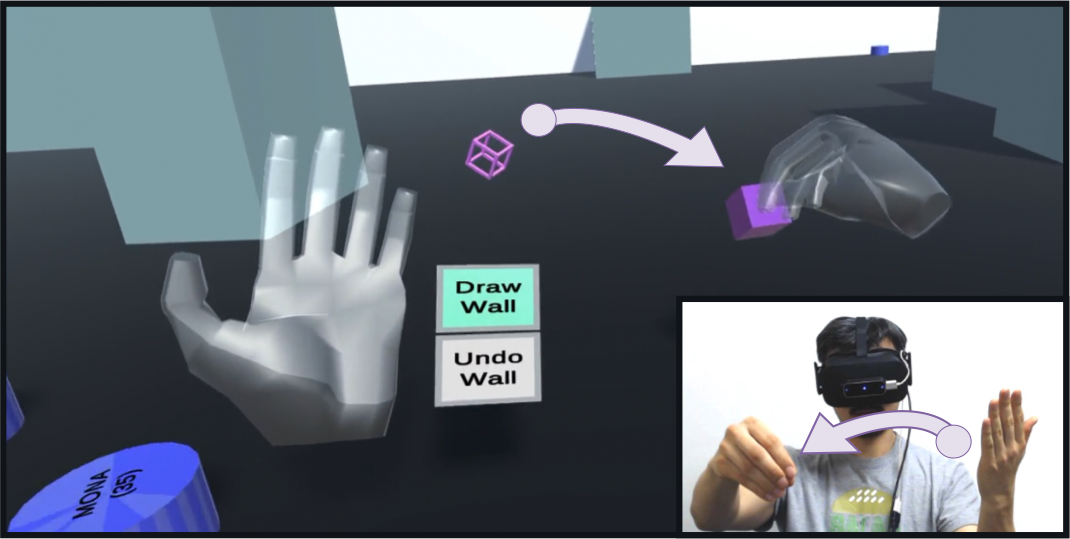}}
\hfil
\caption{(a) The hand-held menu for switching interaction modes; (b) Swarm-oriented interface: multi-robot control cube}\label{fig:UI_menu}
\end{figure}

\textbf{Multi-robot Control Cube:}
The user can have a small hand-held menu by rotating the left-hand palm to face up, as shown in Fig. \ref{fig:UI_menu}(a). 
On the top, there is a pickable cube, which can serve as a virtual guided point of multi-robot coordination, e.g. 
the virtual centre of a rotating formation control\cite{hu2018innovative}. 
In this work, this formation control is activated once the cube is placed on the floor.

\begin{figure}[t]
\includegraphics[width=0.49\linewidth]{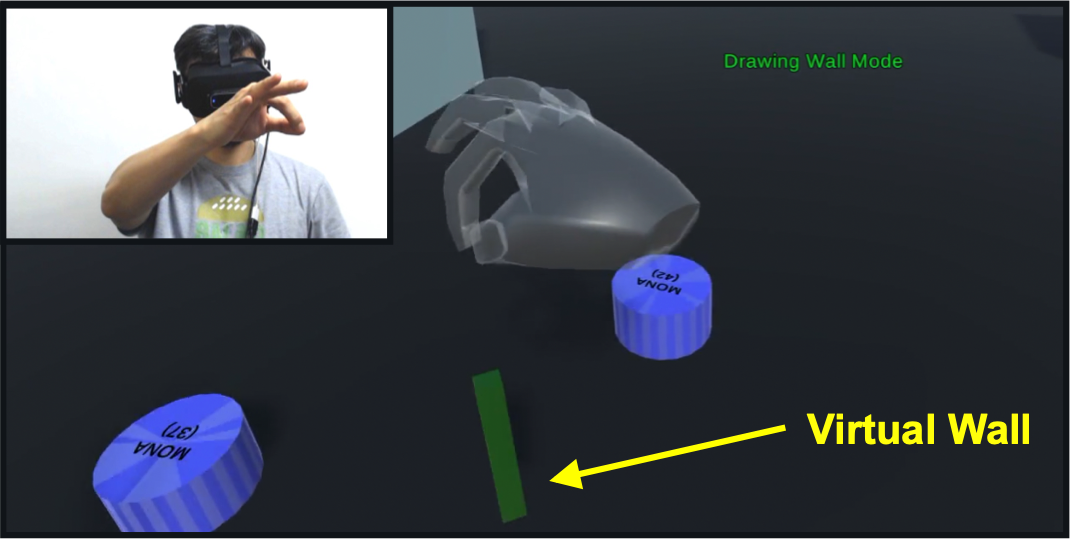}
\includegraphics[width=0.49\linewidth]{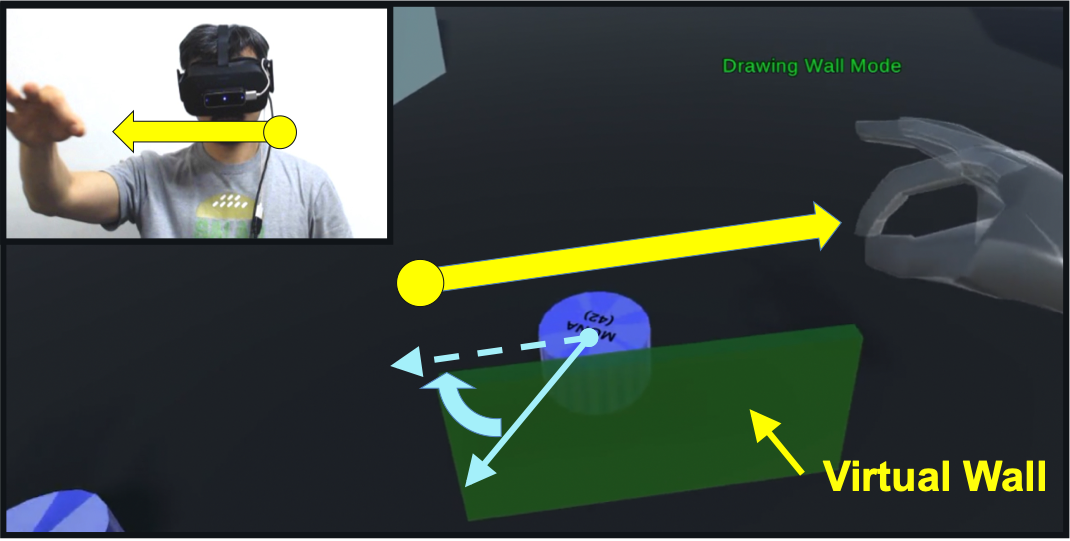}
\caption{Environment-oriented interface: creating a virtual wall}\label{fig:UI_wall}
\end{figure}

\textbf{Virtual Wall:}
In the hand-held menu shown in Fig. \ref{fig:UI_menu}(a), there are two buttons: \emph{Draw Wall} and \emph{Undo Wall}. 
By touching the former, the \emph{drawing wall mode} is toggled on, then a red-coloured sign appears on the VR display. 
In the mode, a pinching gesture creates a linear virtual wall, as shown in Fig. \ref{fig:UI_wall}.  
Such a wall, to which any robot in reality cannot penetrate, indirectly guides the robot's path or confines the robot within a certain area. 
Each wall can be cleared out if the undo wall button is pushed in a last-come-first-served manner.
For its implementation with consideration of reducing communication costs towards real robots, we set that once a wall is created, only its two end positions are broadcasted. Then, the robots compute additional intermediate points depending on their collision avoidance radii, and do collision avoidance behaviours against all the points.


\section{Experimental Analysis}\label{sec:experiments}

\subsection{Experimental Validation using Mixed Reality}

Proof-of-concept experiments to validate the proposed HSI framework use a mixed reality environment where three real MONA robots\cite{Arvin2018} (whose height and diameter is 40 and 74 mm, respectively) and six virtual robots are moving around a $90 \times 150$ cm of arena. 
In the experiments, the robots basically do random walking unless a human operator intervenes their behaviours.
The robots are capable of simple collision avoidance against any of virtual (or real) walls and robots. 
Their localisation relies on a low-cost USB camera-based tracking system\cite{Krajnik2014jint}, which obtains the planar positions and heading angles of the real robots in the arena and sends the information to the master control computer.
The master system consists of a computer executing the implemented Unity application on Windows 10, which renders the virtual world, and another computer running ROS on Ubuntu 16.04, which sends user inputs to the real robots via an antenna. 

\begin{figure}[t]
\centering
\subfloat[]{\includegraphics[width=0.40\linewidth]{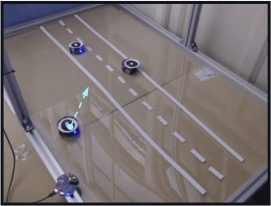} \includegraphics[width=0.40\linewidth]{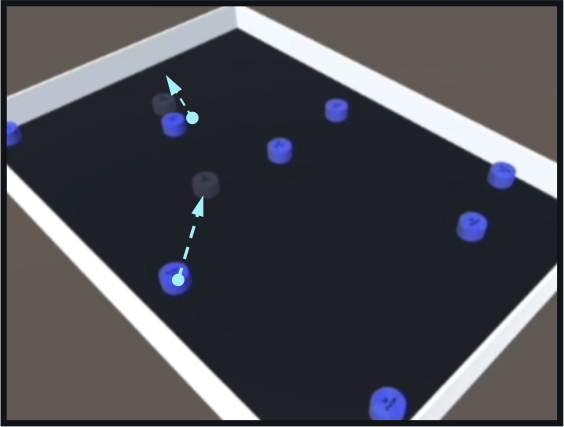}}
\hfil
\subfloat[]{\includegraphics[width=0.40\linewidth]{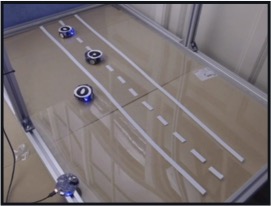} \includegraphics[width=0.40\linewidth]{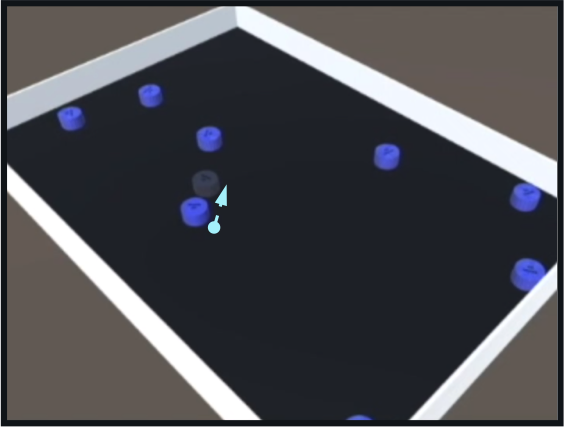}}
\caption{Experimental validation of the pick-and-place interface: (a) once the holographic objects of robots are relocated, (b) the real robots move towards them. The left subfigures show the real robots, and the right subfigures show their visualisation in the virtual space and the other virtual robots. The dashed arrows indicate the remaining journey to the target objects.}  \label{fig:exp_val_individualcontrol}
\end{figure}

\begin{figure}[t]
\centering
\subfloat[]{\includegraphics[width=0.40\linewidth]{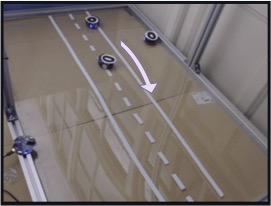} \includegraphics[width=0.40\linewidth]{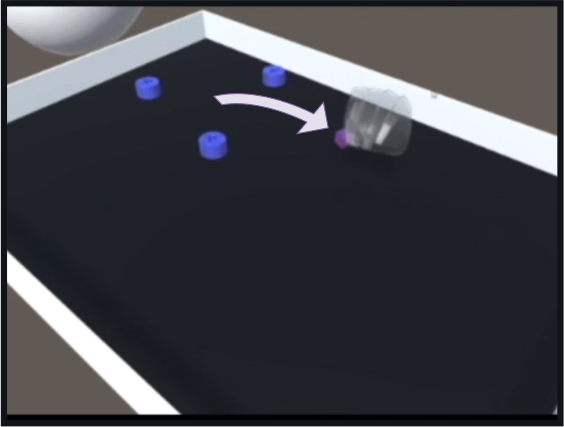}}
\hfil
\subfloat[]{\includegraphics[width=0.40\linewidth]{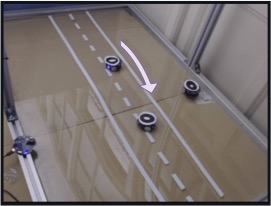} \includegraphics[width=0.40\linewidth]{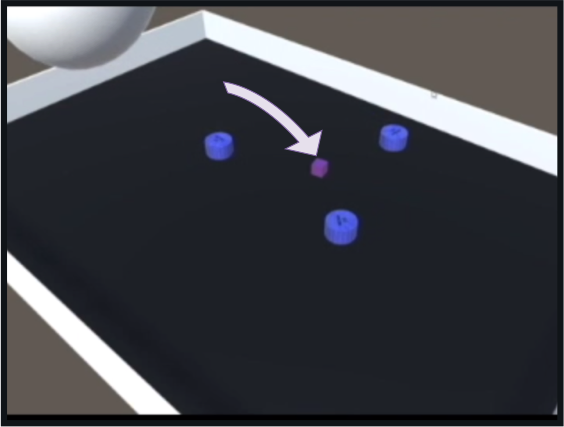}}
\caption{Experimental validation of the multi-robot control cube interface for rotating formation control: (a) once the purple cube object is placed down, (b) the robots forms a circular formation with regard to the cube's position.} 
\label{fig:exp_val_formationcontrol}
\end{figure}

\begin{figure}[t]
\centering
\includegraphics[width=0.40\linewidth]{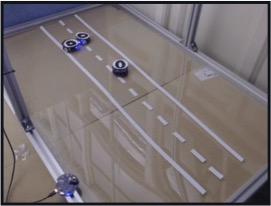} \includegraphics[width=0.40\linewidth]{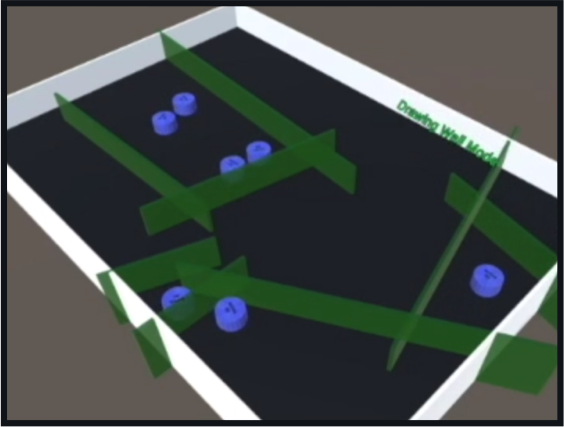}
\caption{Experimental validation of the virtual wall interface: due to the virtual walls (i.e. the green linear objects in the right subfigure), the real or virtual robots are confined within certain spaces.}  
\label{fig:exp_val_virtualwall}
\end{figure}

An experimental demonstration for the pick-and-place interface is presented in Fig. \ref{fig:exp_val_individualcontrol}. 
In the virtual reality, once the target-to-go object of a robot was picked up and placed as in Fig. \ref{fig:exp_val_individualcontrol}(a), the robot moved towards the destination very well. 
Fig. \ref{fig:exp_val_formationcontrol} shows another demonstration for the multi-robot control cube interface. 
In this test, only the real robots were used and the distributed rotating formation control algorithm in \cite{hu2018innovative} was implemented into each of the robots. 
Once a human operator placed down the cube object, the robots started to rotate around it. 
As soon as the cube was relocated as in Fig. \ref{fig:exp_val_formationcontrol}(a), their formation was also changed accordingly as in Fig. \ref{fig:exp_val_formationcontrol}(b). 
A demonstration for the virtual wall interface is shown in Fig. \ref{fig:exp_val_virtualwall}.
Regardless of whether robots are real or virtual, their behaviours were restricted by the virtual walls created by the user. 
All the demonstrations were recorded and can be found in the supplementary material.

%

\subsection{Usability Study}

We conducted a usability analysis to study i) how much the proposed HSI is useful to interact with swarm robots; and ii) how it is effective to be given multiple types of user interfaces. 

\subsubsection{Mission Scenario}
\begin{figure}
\centering
\includegraphics[width=0.80\linewidth]{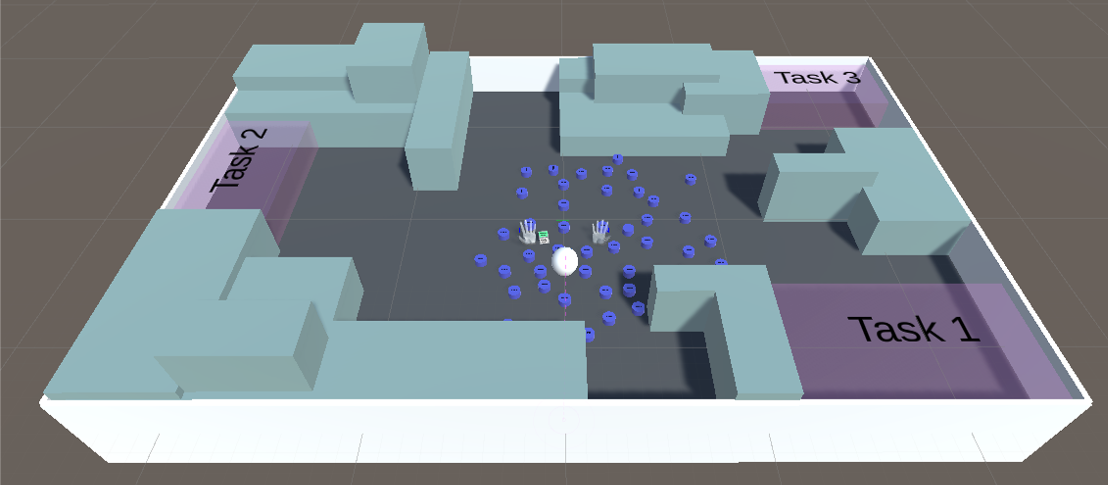}
\caption{The mission arena for the usability study: each participant has to  allocate 50 mobile robots according to the task demands (i.e. 25, 15, and 10 for Task 1, 2, and 3, respectively). The white oval shape represents the user avatar at the time when the mission starts.}
\label{fig:mission}
\end{figure}

The swarm robotic mission that was designed for the usability study is a multi-robot task allocation. 
The objective of the mission is to distribute 50 virtual mobile robots over the three task areas according to their demands (i.e. 25, 15, and 10 robots, respectively), as shown in Fig. \ref{fig:mission}.
The local behaviour of each robot was designed to move forward until it faces an obstacle, then the robot performs a collision avoidance routine by rotating randomly. 
The simplified intelligence would require a human operator's intervention to address the given mission efficiently. 
For this test, all the perception and control interfaces in Sec. \ref{sec:proposed_interface} were used except the one for formation control.




\subsubsection{Experimental Setup}
We recruited 10 participants aged between 20 and 35 from the engineering discipline.  
Half of them had a little experience of VR, and the other half had no experience at all. Since all of them had never used LM before, they were given a five-minute trial of an introductory application called \emph{Blocks}\footnote{\href{https://gallery.leapmotion.com/blocks/}{https://gallery.leapmotion.com/blocks/}} before starting the main test. 
Hence, the mission scenario and the user interfaces they can use were explained.

Each participant was provided two strategies to address the mission. 
In Strategy 1, for controlling the robots, the pick-and-place interface was only allowed to use. 
In Strategy 2, the participants can also use the virtual wall interface.
We explained to them that virtual walls are supposed to block any task arena that already has the required number of robots to prevent it from including any redundant robots. 
Otherwise, the possible results would be affected by the individuals' preferred approaches to address the mission. 
All the perception interfaces were used for both strategies.

Each participant performed the mission using the two strategies respectively, for each of which, two trials were given. 
The trial minimising the completion time was chosen as his/her best performance, and the completion time and the number of interactions they used were recorded.
The participants were also asked to fill a Likert scale survey form to quantify their experience on the individual interfaces as well as the overall system.

\subsubsection{Results and Discussion}

\begin{table}
\caption{Average performance}
\begin{center}
\begin{tabular}{|c||c|c|c|c|}
    \hline
    \multirow{2}{*}{} & \multicolumn{2}{c|}{Strategy 1 (PP)} & \multicolumn{2}{c|}{Strategy 2(PP+VW)} \\
    \cline{2-5}
                     & Ave & Std & Ave & Std \\
    \hline \hline
    Completion time (sec)  & 269.9 & 60.8 & 312.8 & 70.6 \\
    \hline
    The number of interactions  & 27.5  & 6.8  & 21.4 & 6.9 \\
    \hline
    \multicolumn{5}{l}{PP: Pick-and-Place interface;  VW: Virtual Wall interface}
\end{tabular}
\end{center}
\label{tab:performance_result}
\end{table}

\begin{figure}
\centering
\includegraphics[width=0.95\linewidth]{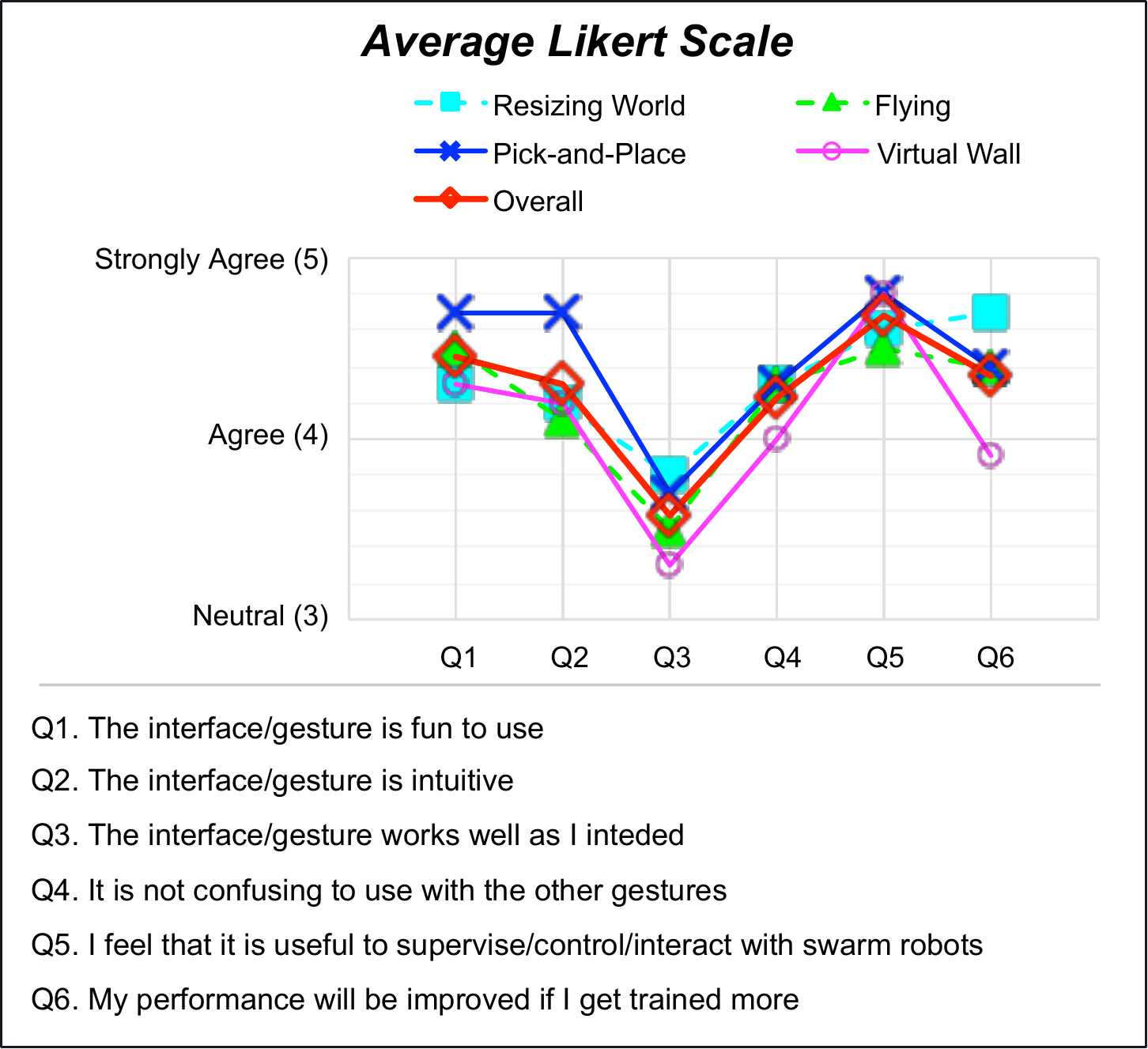}
\caption{Qualitative Comparative Result of HSI}
\label{fig:result_usability}
\end{figure}

Table \ref{tab:performance_result} shows that Strategy 1 (i.e. the pick-and-place interface only in use) averagely requires less time (i.e. 43 sec less) but more interactions (i.e. 6.1 interactions more), compared with Strategy 2 (i.e. virtual walls also in use). 
This indicates that using environment-oriented controls can reduce the needs of explicit one-by-one guidance towards individual robots, ending up with reduction in the total number of interactions.
On the contrary, the increase in the completion time implies that a user may be confused with multiple modalities, especially, when the interfaces are similar to each other. 
Even this result was also the case for experienced users (i.e., the developers of the proposed system) because toggling on and off the drawing wall mode increases a mission completion time. 

Fig. \ref{fig:result_usability} presents the user experience result of the proposed system. 
The average answers for Q3 imply that users may need to get more trained to use LM. In fact, during the test, it was often observed that the participants unconsciously stretched out their hands out of LM's sensing range. 
The result can be considered obvious due to the fact that the end-user interface with VR and LM was definitely unfamiliar to the participants.
The answers for Q6 indicate that the resizing world interface relatively needs more training, whereas the virtual wall interface is easier to use. 
In contrast, the virtual wall interface was selected as the most confusing one, as in the results for Q4. This seems to be relevant to the increased completion time in Stratege 2 as in Table \ref{tab:performance_result}, because the pinching gesture is used to create virtual walls as well as to resize the world, but in different toggle modes, respectively.   

However, it was mostly agreed that the proposed HSI framework would be useful for interaction with swarm robots, as in the result for Q5.
Obviously, the pick-and-place interface is the most fun and intuitive according to the results for Q1 and Q2.



\section{Conclusions}\label{sec:conclusion}

This paper proposed an intuitive human-swarm interaction framework using a super-powered user avatar who can supervise and interact with a swarm of mobile robots in virtual reality, which is implemented by VR and Leap Motion. 
This work presented two perception interfaces, by which a user can resize the virtual world or fly around the scene, and three control interfaces for robot-oriented, swarm-oriented, and environment-oriented interactions, respectively. 
We conducted proof-of-concept experiments to validate the proposed HSI framework by using three real robots, MONA, and six virtual ones in a mixed reality environment. 
A usability study for a multi-robot task allocation mission was used to evaluate  the proposed framework. 
The results presented that the proposed system can be considered as suitable for swarm robots in a vast and remote environment, and that the individual interfaces using bare hands are intuitive. 
It was also shown that multiple modalities can reduce the number of human intervention, but may increase a mission completion time, especially if users are not trained enough, due to its inherent complexity.


For real world application, 
the communication capability of swarm robots to a human operator will be one of the big challenges. 
Considering any possible practical network topology, 
the larger number will impose huge communication load on the near-end robots as well as cause bottleneck effects on the information flow. 
Eventually, this will lead to a latency of remote visualisation for the operator. 
Therefore, the near-end robots or any robots in the middle may need to make decisions in terms of which information from which robots needs to be priorly transferred in order to maximise the operator's perception, while reducing communication load imposed on the near-end robots. 

\bibliographystyle{IEEEtran}
\bibliography{library,FA}

\begin{thebibliography}{10}
\providecommand{\url}[1]{#1}
\csname url@rmstyle\endcsname
\providecommand{\newblock}{\relax}
\providecommand{\bibinfo}[2]{#2}
\providecommand\BIBentrySTDinterwordspacing{\spaceskip=0pt\relax}
\providecommand\BIBentryALTinterwordstretchfactor{4}
\providecommand\BIBentryALTinterwordspacing{\spaceskip=\fontdimen2\font plus
\BIBentryALTinterwordstretchfactor\fontdimen3\font minus
  \fontdimen4\font\relax}
\providecommand\BIBforeignlanguage[2]{{%
\expandafter\ifx\csname l@#1\endcsname\relax
\typeout{** WARNING: IEEEtran.bst: No hyphenation pattern has been}%
\typeout{** loaded for the language `#1'. Using the pattern for}%
\typeout{** the default language instead.}%
\else
\language=\csname l@#1\endcsname
\fi
#2}}

\bibitem{hamann2018swarm}
H.~Hamann, \emph{Swarm Robotics: A Formal Approach}.\hskip 1em plus 0.5em minus
  0.4em\relax Springer, 2018.

\bibitem{Trianni2018}
V.~Trianni, M.~Trabattoni, A.~Antoun, E.~Hocquard, B.~Wiandt, G.~Valentini,
  M.~Dorigo, and Y.~Tamura, ``{Kilogrid: a novel experimental environment for
  the Kilobot robot},'' \emph{Swarm Intelligence}, vol.~12, no.~3, pp.
  245--266, 2018.

\bibitem{Arvin2018}
F.~Arvin, J.~Espinosa, B.~Bird, A.~West, S.~Watson, and B.~Lennox, ``{Mona: an
  affordable open-source mobile robot for education and research},''
  \emph{Journal of Intelligent and Robotic Systems},
  pp. 1--15, 2018.

\bibitem{Nagi2015}
J.~Nagi, H.~Ngo, L.~M. Gambardella, and G.~A. {Di Caro}, ``{Wisdom of the swarm
  for cooperative decision-making in human-swarm interaction},'' in \emph{IEEE
  Intl. Conf. on Robotics and Automation}, 2015, pp. 1802--1808.

\bibitem{Choi2009}
H.~L. Choi, L.~Brunet, and J.~P. How, ``{Consensus-based decentralized auctions
  for robust task allocation},'' \emph{IEEE Transactions on Robotics}, vol.~25,
  no.~4, pp. 912--926, 2009.

\bibitem{Jang2018a}
I.~Jang, H.-S. Shin, and A.~Tsourdos, ``{Anonymous hedonic game for task
  allocation in a large-scale multiple agent system},'' \emph{IEEE Transactions
  on Robotics}, vol.~34, no.~6, pp. 1534--1548, 2018.

\bibitem{Jang2018d}
I.~Jang, H.-S. Shin, and A.~Tsourdos, ``{Local information-based control for
  probabilistic swarm distribution guidance},'' \emph{Swarm Intelligence},
  vol.~12, no.~4, pp. 327--359, 2018.

\bibitem{Nagi2014}
J.~Nagi, A.~Giusti, L.~M. Gambardella, and G.~A. {Di Caro}, ``{Human-swarm
  interaction using spatial gestures},'' in \emph{IEEE/RSJ Intl. Conf. on Intelligent Robots and Systems}, 2014, pp. 3834--3841.

\bibitem{Podevijn2014}
G.~Podevijn, R.~O'Grady, Y.~S.~G. Nashed, and M.~Dorigo, ``{Gesturing at
  Subswarms: Towards Direct Human Control of Robot Swarms},'' in \emph{Towards
  Autonomous Robotic Systems 2013. LNCS}, pp. 390--403.

\bibitem{Alonso-Mora2015}
J.~Alonso-Mora, S.~{Haegeli Lohaus}, P.~Leemann, R.~Siegwart, and P.~Beardsley,
  ``{Gesture based human-multi-robot swarm interaction and its application to
  an interactive display},'' in \emph{IEEE Intl. Conf. on Robotics
  and Automation}, 2015, pp.
  5948--5953.

\bibitem{Stoica2013}
A.~Stoica, T.~Theodoridis, H.~Hu, K.~McDonald-Maier, and D.~F. Barrero,
  ``{Towards human-friendly efficient control of multi-robot teams},'' in
  \emph{Intl. Conf. on Collaboration Technologies and
  Systems}, 2013, pp. 226--231.

\bibitem{Gromov2016}
B.~Gromov, L.~M. Gambardella, and G.~A. {Di Caro}, ``{Wearable multi-modal
  interface for human multi-robot interaction},'' in \emph{Intl. Symp. on Safety, Security and Rescue Robotics}, 2016, pp. 240--245.

\bibitem{Frank2017}
J.~A. Frank, S.~P. Krishnamoorthy, and V.~Kapila, ``{Toward mobile
  mixed-reality interaction with multi-robot systems},'' \emph{IEEE Robotics
  and Automation Letters}, vol.~2, no.~4, pp. 1901--1908, 2017.

\bibitem{Patel2019}
\BIBentryALTinterwordspacing
J.~Patel, Y.~Xu, and C.~Pinciroli, ``{Mixed-Granularity Human-Swarm
  Interaction},'' in \emph{Intl. Conf. on Robotics Automation 2019 (in
  press)}. [Online]. Available: \url{http://arxiv.org/abs/1901.08522}
\BIBentrySTDinterwordspacing

\bibitem{LeGoc2016}
M.~{Le Goc}, L.~H. Kim, A.~Parsaei, J.-D. Fekete, P.~Dragicevic, and
  S.~Follmer, ``{Zooids: building blocks for swarm user interfaces},'' in
  \emph{Proc. of the 29th Annual Symp. on User Interface Software and
  Technology}, 2016, pp. 97--109.

\bibitem{Roldan2019}
J.~J. Rold{\'{a}}n, E.~Pe{\~{n}}a-Tapia, D.~Garz{\'{o}}n-Ramos,
  J.~de~Le{\'{o}}n, M.~Garz{\'{o}}n, J.~del Cerro, and A.~Barrientos,
  ``{Multi-robot Systems, virtual reality and ROS: developing a new generation
  of operator interfaces},'' in \emph{Robot Operating System (ROS), Studies in
  Computational Intelligence}.\hskip 1em plus 0.5em minus 0.4em\relax Springer
  International Publishing, 2019, vol. 778, pp. 29--64.

\bibitem{Pinciroli2011}
C.~Pinciroli, V.~Trianni, and R.~O'Grady, ``{ARGoS: a modular, multi-engine
  simulator for heterogeneous swarm robotics},'' in \emph{IEEE/RSJ
  Intl. Conf. on Intelligent Robots and Systems}, 2011, pp. 5027--5034.

\bibitem{Whitney2018}
D.~Whitney, E.~Rosen, D.~Ullman, E.~Phillips, and S.~Tellex, ``{ROS reality: a
  virtual reality framework using consumer-grade hardware for ROS-enabled
  robots},'' in \emph{IEEE/RSJ Intl. Conf. on Intelligent Robots
  and Systems}, 2018.

\bibitem{Jang2019}
I.~Jang, J.~Carrasco, A.~Weightman, and B.~Lennox, ``{Intuitive bare-hand
  teleoperation of a robotic manipulator using virtual reality and leap
  motion},'' in \emph{Towards Autonomous Robotic Systems 2019 (submitted)}.

\bibitem{Jang2018}
I.~Jang, H.-S. Shin, A.~Tsourdos, J.~Jeong, S.~Kim, and J.~Suk, ``{An
  integrated decision-making framework of a heterogeneous aerial robotic swarm
  for cooperative tasks with minimum requirements},'' \emph{Proceedings of the
  Institution of Mechanical Engineers, Part G: Journal of Aerospace
  Engineering}, 2018.

\bibitem{Bird2018}
B.~Bird, A.~Griffiths, H.~Martin, E.~Codres, J.~Jones, A.~Stancu, B.~Lennox,
  S.~Watson, and X.~Poteau, ``{Radiological monitoring of nuclear facilities:
  using the continuous autonomous radiation monitoring assistance robot},''
  \emph{IEEE Robotics {\&} Automation Magazine}, 2018.

\bibitem{martinoli-04}
A.~Martinoli, K.~Easton, and W.~Agassounon, ``Modeling swarm robotic systems: A
  case study in collaborative distributed manipulation,'' \emph{The
  International Journal of Robotics Research}, vol.~23, no. 4-5, pp. 415--436,
  2004.

\bibitem{Arvin2015}
F.~Arvin, T.~Krajnik, A.~E. Turgut, and S.~Yue, ``{COS$\Phi$: Artificial
  pheromone system for robotic swarms research},'' in \emph{IEEE/RSJ Intl. Conf. on Intelligent Robots and Systems}, 2015, pp. 407--412.


\bibitem{Krajnik2014jint}
T.~Krajn{\' i}k, M.~Nitsche, J.~Faigl, P.~Van{\v e}k, M.~Saska, L.~ P{\v r}eu{\v c}il, T.~Duckett, and M.~Mejail, ``A Practical Multirobot Localization System,'' \emph{Journal of Intelligent \& Robotic Systems}, vol.~76, no. 3-4, pp. 539--562, 2014.


\bibitem{hu2018innovative}
J.~Hu and A.~Lanzon, ``An innovative tri-rotor drone and associated distributed
  aerial drone swarm control,'' \emph{Robotics and Autonomous Systems}, vol.
  103, pp. 162--174, 2018.



\end{thebibliography}

\end{document}